\documentclass[letterpaper, 10 pt, conference]{ieeeconf}
\IEEEoverridecommandlockouts
\usepackage{cite}
\usepackage{amsmath,amssymb,amsfonts}
\usepackage{dsfont}
\usepackage{mathtools}
\usepackage{algorithm}
\usepackage{algpseudocode}

\usepackage{graphicx}
\usepackage{textcomp}
\usepackage{xcolor}

\usepackage{tikz}
\usepackage{pgfplots} 
\pgfplotsset{compat=newest}
\usetikzlibrary{spy, arrows, positioning, shapes, shapes.geometric, external, shadows, babel, calc, backgrounds, intersections, angles}

\usepackage{booktabs}
\usepackage{diagbox}
\usepackage{siunitx}
\usepackage[labelformat=simple]{subcaption}

\usepackage{makecell}
\makeatletter
\let\NAT@parse\undefined
\makeatother
\usepackage[hidelinks]{hyperref}
\usepackage{cleveref}

\DeclarePairedDelimiter{\norm}{\lVert}{\rVert}

\definecolor{tpGreenColor}{HTML}{028F1E}
\definecolor{tnGreenColor}{HTML}{21FC0D}
\definecolor{fnRedColor}{HTML}{9E003A}
\definecolor{ambiguousColor}{HTML}{FEC615}
\definecolor{greyColor}{HTML}{59656D}

\pgfplotsset{
	colormap={cm_radar_dl_gt}{color(0)=(greyColor), color(1)=(greyColor), color(2)=(blue), color(3)=(red), color(4)=(ambiguousColor)}
}

\pgfplotsset{
	colormap={cm_radar_dl_confusion}{color(0)=(greyColor), color(1)=(tpGreenColor), color(2)=(red), color(3)=(tnGreenColor), color(4)=(fnRedColor), color(5)=(ambiguousColor)}
}

\newcommand{\plotRadarDetectionListGT}[1] {
	\addplot
	[
		scatter,
		only marks,
		scatter src=explicit symbolic,
		scatter/classes={
			0={mark=*, lime},
			1={mark=*, mark size=1.0pt, greyColor},
			2={mark=*, blue},
			3={mark=*, red},
			4={mark=*, ambiguousColor}
		},
		mark size=1.25pt
	]
	table
	[
		x expr=\thisrow{distance} * sin(\thisrow{azimuth} * 180 / pi),
		y expr=\thisrow{distance} * cos(\thisrow{azimuth} * 180 / pi),
		meta=gt_label,
		col sep=comma
	] {#1};
	
	\addplot
	[
		point meta=explicit symbolic,
		quiver={
			u=\thisrow{absolute_velocity} * sin(\thisrow{azimuth} * 180 / pi),
			v=\thisrow{absolute_velocity} * cos(\thisrow{azimuth} * 180 / pi),
			scale arrows=1.0,
			colored
		},
		colormap access=direct,
		-stealth
	]
	table
	[
		x expr=\thisrow{distance} * sin(\thisrow{azimuth} * 180 / pi),
		y expr=\thisrow{distance} * cos(\thisrow{azimuth} * 180 / pi),
		meta=gt_label,
		col sep=comma,
		restrict expr to domain={abs(\thisrow{absolute_velocity})}{1.5:+inf}
	] {#1};
}

\newcommand{\plotRadarDetectionListConfusion}[1] {
	\addplot
	[
		scatter,
		only marks,
		scatter src=explicit symbolic,
		scatter/classes={
			0={mark=*, mark size=1.0pt, greyColor},
			1={mark=*, tpGreenColor},
			2={mark=*, red},
			3={mark=*, tnGreenColor},
			4={mark=*, fnRedColor},
			5={mark=*, ambiguousColor}
		},
		mark size=1.25pt
	]
	table
	[
		x expr=\thisrow{distance} * sin(\thisrow{azimuth} * 180 / pi),
		y expr=\thisrow{distance} * cos(\thisrow{azimuth} * 180 / pi),
		meta=confusion_value,
		col sep=comma
	] {#1};
	
	\addplot
	[
		point meta=explicit symbolic,
		quiver={
			u=\thisrow{absolute_velocity} * sin(\thisrow{azimuth} * 180 / pi),
			v=\thisrow{absolute_velocity} * cos(\thisrow{azimuth} * 180 / pi),
			scale arrows=1.0,
			colored
		},
		colormap access=direct,
		-stealth
	]
	table
	[
		x expr=\thisrow{distance} * sin(\thisrow{azimuth} * 180 / pi),
		y expr=\thisrow{distance} * cos(\thisrow{azimuth} * 180 / pi),
		meta=confusion_value,
		col sep=comma,
		restrict expr to domain={abs(\thisrow{absolute_velocity})}{1.5:+inf}
	] {#1};
}

\newcommand{\drawObject}[7] {
	\draw[cyan!70!white, fill] (axis cs: {#2 + #4 * sin(#3) + #6 * cos(#3)}, {#1 + #4 * cos(#3) - #6 * sin(#3)}) -- (axis cs: {#2 + #4 * sin(#3) + #7 * cos(#3)}, {#1 + #4 * cos(#3) - #7 * sin(#3)}) -- (axis cs: {#2 + #5 * sin(#3) + #7 * cos(#3)}, {#1 + #5 * cos(#3) - #7 * sin(#3)}) -- (axis cs: {#2 + #5 * sin(#3) + #6 * cos(#3)}, {#1 + #5 * cos(#3) - #6 * sin(#3)}) -- cycle;
}

\title{\LARGE \textbf{Fast Rule-Based Clutter Detection in Automotive Radar Data}}

\author{Johannes Kopp\textsuperscript{1}, Dominik Kellner\textsuperscript{2}, Aldi Piroli\textsuperscript{1} and Klaus Dietmayer\textsuperscript{1}
\thanks{\textsuperscript{1}Institute of Measurement, Control and Microtechnology, Ulm University, Albert-Einstein-Allee 41, 89081 Ulm, Germany {\tt\small \{firstname\}.\{lastname\}@uni-ulm.de}}%
\thanks{\textsuperscript{2}BMW AG, Petuelring 130, 80809 Munich, Germany {\tt\small dominik.m.kellner@bmw.de}}%
}

\newcommand\copyrighttext{
	\footnotesize \textcopyright 2021 IEEE.  Personal use of this material is permitted.  Permission from IEEE must be obtained for all other uses, in any current or future media, including reprinting/republishing this material for advertising or promotional purposes, creating new collective works, for resale or redistribution to servers or lists, or reuse of any copyrighted component of this work in other works.}
\newcommand\copyrightnotice[1]{
	\begin{tikzpicture}[remember picture,overlay]
		\node[anchor=north,yshift=-15pt] at (current page.north) {\parbox{\dimexpr\textwidth-1.0cm}{#1}};
	\end{tikzpicture}
	\vspace{-10pt}
}

\begin{document}

\maketitle
\copyrightnotice{\copyrighttext}
\thispagestyle{empty}
\pagestyle{empty}

\begin{abstract}

Automotive radar sensors output a lot of unwanted clutter or ghost detections, whose position and velocity do not correspond to any real object in the sensor's field of view. This poses a substantial challenge for environment perception methods like object detection or tracking. Especially problematic are clutter detections that occur in groups or at similar locations in multiple consecutive measurements. In this paper, a new algorithm for identifying such erroneous detections is presented. It is mainly based on the modeling of specific commonly occurring wave propagation paths that lead to clutter. In particular, the three effects explicitly covered are reflections at the underbody of a car or truck, signals traveling back and forth between the vehicle on which the sensor is mounted and another object, and multipath propagation via specular reflection. The latter often occurs near guardrails, concrete walls or similar reflective surfaces. Each of these effects is described both theoretically and regarding a method for identifying the corresponding clutter detections. Identification is done by analyzing detections generated from a single sensor measurement only. The final algorithm is evaluated on recordings of real extra-urban traffic. For labeling, a semi-automatic process is employed. The results are promising, both in terms of performance and regarding the very low execution time. Typically, a large part of clutter is found, while only a small ratio of detections corresponding to real objects are falsely classified by the algorithm.

\end{abstract}

\section{Introduction}

\begin{figure}[!t]
		\hspace*{-0.1cm}
		\begin{tikzpicture}
		[
			every axis/.append style = {label style={font=\footnotesize}, tick label style={font=\footnotesize}},
			every pin/.style = {font=\footnotesize, align=center},
			every pin edge/.style = {black, densely dashed}
		]
		\begin{axis}
		[
			width=1.20*\columnwidth,
			grid=major,
			grid style={dashed,gray!30},
			xlabel= $y$ (\si{\meter}),
			ylabel= $x$ (\si{\meter}),
			x label style={yshift=0.15cm},
			x dir=reverse,
			y label style={yshift=-0.25cm},
			axis equal image,
			xmin=-50,
			xmax=50,
			ymax=100,
			ymin=0,
			colormap name=cm_radar_dl_gt
		]
		
			\drawObject{52.200793227351255}{-0.9938576201745946}{-1.8798025420001778}{-0.2}{-4.2}{1.0}{-1.0}
			\drawObject{68.71518440817454}{15.791115263230267}{179.2757036388099}{2.3}{-2.3}{1.0}{-1.0}
			\drawObject{69.88644591767513}{11.855546439145883}{179.49469017219985}{2.0}{-4.0}{1.25}{-1.25}
			\drawObject{94.18646582261442}{7.23804598068773}{179.0}{2.12}{-2.12}{1.0}{-1.0}
			\drawObject{97.95252970821866}{13.567980927024792}{178.8976706658552}{2.12}{-2.12}{1.0}{-1.0}
			\drawObject{99.83184245131838}{-2.6}{-2.1488226140260283}{2.12}{-2.12}{1.0625}{-1.0625}
			
			\plotRadarDetectionListGT{resources/example_specular_reflection_dl.csv}
			
			\node[inner sep=0pt, outer sep=0pt, anchor=north east] at (axis cs: -50, 100) {\includegraphics[width=3.2cm]{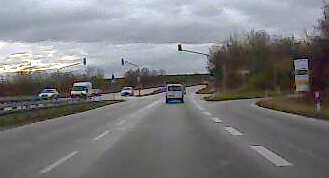}};
			
			\node[rectangle, draw=black, densely dashed, minimum width=0.4cm, minimum height=0.4cm, inner sep=0pt, outer sep=1pt, font=\footnotesize, pin={[fill=white, inner sep=1pt, outer sep=0pt, anchor=center, pin distance=2.12cm, pin edge={shorten >= -0.05cm}] -120.0:Clutter caused by\\specular reflection}] at (axis cs: 12.6, 49.1) {};
		
		\end{axis}
		\end{tikzpicture}

	\caption{Example of a radar sensor's output containing clutter. Points mark the positions of detections relative to the sensor, attached arrows visualize their velocity over ground. Colors indicate what detections correspond to: Reflections at stationary objects or ground are gray, reflections at vehicles blue, clutter is red and ambiguous detections are yellow. For reference, the positions of vehicles are marked with boxes.}
	\label{fig:example_specular_reflection}
\end{figure}

Advanced driver assistance systems and autonomous driving are areas of high importance and continuous development in the automotive industry. In both of them, reliable perception of the vehicle's surroundings is the basis for all high-level processing steps. This includes the detection of road users like vehicles or pedestrians, tracking their movement over time and making decisions like planning the own behavior and trajectory. The sensors typically used for this task are camera, lidar and radar. A major advantage of radar sensors is their robustness to adverse weather conditions, such as rain, fog and snow. Furthermore, they allow for a direct measurement of an object's radial velocity component and have relatively low production cost.

Radar sensors work by emitting electromagnetic waves which get reflected at obstacles in their path and receiving the resulting echoes. From these, individual points of reflection are determined, in the following called detections. For each detection, the distance, angle of arrival and relative velocity in radial direction are estimated. However, many of the detections output at the end of a measurement cycle are ghost targets or clutter, i.e. they don't correspond to any real object within the sensor's field of view (FoV), see e.g. Fig.~\ref{fig:example_specular_reflection}. Section~\ref{section:clutter_effects} describes common causes for this phenomenon.

The high portion of clutter poses a substantial challenge for environment perception methods (exclusively) relying on radar data. Especially problematic are groups of clutter detections occurring consistently over consecutive measurements. For example, a neural network that processes radar detections to detect road users (see e.g.~\cite{Danzer2019}) could falsely recognize such clusters as objects. This in turn might influence driving behavior of an autonomous vehicle utilizing its outputs. Besides trying to further increase the network's robustness to clutter, the described problem can be tackled by employing a dedicated preprocessing step. Its task is to identify and filter as much clutter as possible while ideally preserving all other detections and introducing only minimal time delay. In this work, a rule-based algorithm that realizes this functionality is presented.

\section{Related Work} \label{section:related_work}

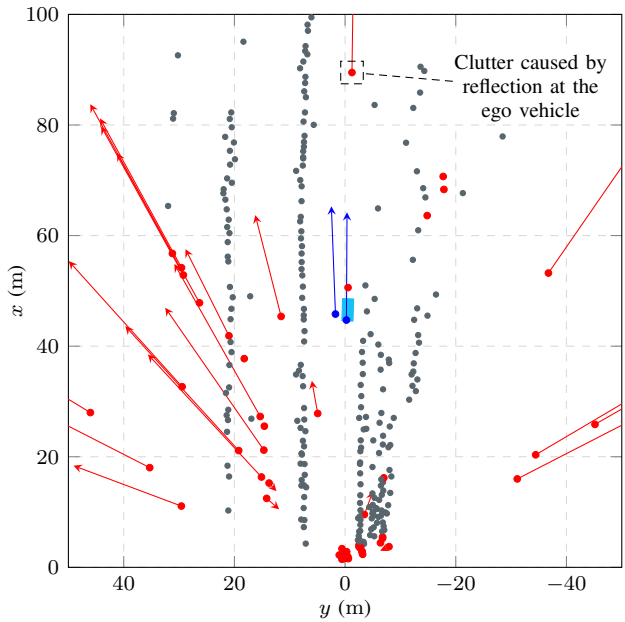
\begin{figure}[!t]
	\hspace*{-0.1cm}
	\begin{tikzpicture}
	[
		every axis/.append style = {label style={font=\footnotesize}, tick label style={font=\footnotesize}},
		every pin/.style = {font=\footnotesize, align=center},
		every pin edge/.style = {black, densely dashed}
	]
	\begin{axis}
	[
		width=1.20*\columnwidth,
		grid=major,
		grid style={dashed,gray!30},
		xlabel= $y$ (\si{\meter}),
		ylabel= $x$ (\si{\meter}),
		x label style={yshift=0.15cm},
		x dir=reverse,
		y label style={yshift=-0.25cm},
		axis equal image,
		xmin=-50,
		xmax=50,
		ymax=100,
		ymin=0,
		colormap name=cm_radar_dl_gt
	]
	
		\drawObject{46.56348510567693}{-0.4847019586679835}{-1.6733408252683306}{2.0}{-2.0}{1.0}{-1.0}
		\drawObject{126.51652655899352}{-5.185580367126022}{-3.9603474743548395}{4.087500095367432}{-4.087500095367432}{1.537500023841858}{-1.537500023841858}
		
		\plotRadarDetectionListGT{resources/example_reflections_at_ego_vehicle_dl.csv}
		
		\node[rectangle, draw=black, densely dashed, minimum width=0.3cm, minimum height=0.3cm, inner sep=0pt, outer sep=1pt, font=\footnotesize, pin={[fill=white, inner sep=1pt, outer sep=0pt, anchor=center, pin distance=2.2cm, pin edge={shorten >= -0.03cm}] -5.0:Clutter caused by\\reflection at the\\ego vehicle}] at (axis cs: -1.25, 89.49) {};
	
	\end{axis}
	\end{tikzpicture}
	
	\caption{Example in which the transmitted signal bounces between a preceding car and the ego vehicle. This leads to a clutter detection in double the distance. Coloring as in Fig.~\ref{fig:example_specular_reflection}.}
	\label{fig:example_reflections_at_ego_vehicle}
\end{figure}

Only few publications explicitly deal with the task of detecting clutter among radar detections. What is more, some of them simplify the problem by considering only a subset of possible clutter, e.g. that resulting from certain effects, or by imposing restrictions during data acquisition and evaluation.

Model- and rule-based approaches are mostly neglected in literature. To the best of the authors' knowledge, the only suitable work in the field of automotive applications is~\cite{Roos2017}. There, detections are segmented into clusters which are presumed to represent vehicles. The motion state of each vehicle is then estimated from the included detections and compared to the orientation of a fitted 2D box. A mismatch between the two indicates a cluster of clutter detections. However, the method requires two radar sensors with overlapping FoV that both measure multiple detections per object. Furthermore, only clutter caused by certain types of specular reflection can be detected this way.

On the side of machine learning, multiple relevant publications exist.
In~\cite{Prophet2019}, several hand-crafted features are calculated for each moving detection and appended to the values measured by the sensor. The result is then fed to a classifier which tries to differentiate between nonclutter and two types of clutter detections. While multiple models are tested, the best performance is achieved using a random forest~\cite{Breiman2001}.
The same authors explore a different approach in~\cite{Garcia2019}. Here, a convolutional neural network is employed. Its inputs are a 2D map of all moving detections in a measurement and an occupancy grid map generated from detections accumulated over time. The resulting heat map is utilized to identify clutter. However, only the output for detections very close to the sensor is actually evaluated.
In~\cite{Kraus2020}, the modified version of PointNet++ that was introduced in~\cite{Schumann2018} is used to distinguish between detections stemming from road users, clutter resulting from a particular type of specular reflection and other detections (among other class configurations). Detections of two experimental radar sensors accumulated over \SI{200}{ms} serve as input to the network. The used data set, while having considerable size, consists only of very specific scenarios recorded not in real traffic but in a controlled environment and with stationary sensors.
In~\cite{Chamseddine2021}, it is attempted to find clutter within the detections of a high-resolution 3D radar sensor. To this end, PointNet~\cite{PointNet} is adapted to better support the specific input features provided for each detection. Furthermore, a method for automatic labeling based on comparing spatial positions of radar and lidar points in image space is presented.

\section{Clutter Effects} \label{section:clutter_effects}

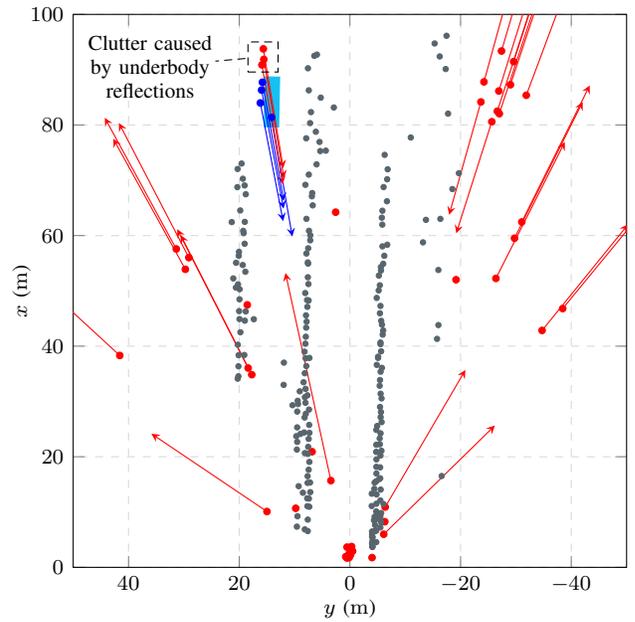
\begin{figure}[!t]
	\hspace*{-0.1cm}
	\begin{tikzpicture}
	[
		every axis/.append style = {label style={font=\footnotesize}, tick label style={font=\footnotesize}},
		every pin/.style = {font=\footnotesize, align=center},
		every pin edge/.style = {black, densely dashed}
	]
	\begin{axis}
	[
		width=1.20*\columnwidth,
		grid=major,
		grid style={dashed,gray!30},
		xlabel= $y$ (\si{\meter}),
		ylabel= $x$ (\si{\meter}),
		x label style={yshift=0.15cm},
		x dir=reverse,
		y label style={yshift=-0.25cm},
		axis equal image,
		xmin=-50,
		xmax=50,
		ymax=100,
		ymin=0,
		colormap name=cm_radar_dl_gt
	]
	
		\drawObject{83.68134424024436}{14.096845570640994}{178.86054190898696}{4.0}{-5.0}{1.3}{-1.3}
		
		\plotRadarDetectionListGT{resources/example_underbody_reflections_dl.csv}
		
		\node[rectangle, draw=black, densely dashed, minimum width=0.4cm, minimum height=0.4cm, inner sep=0pt, outer sep=1pt, font=\footnotesize, pin={[fill=white, inner sep=1pt, outer sep=0pt, anchor=center, pin distance=1.27cm, pin edge={shorten >= 0.01cm}] -175.0:Clutter caused\\by underbody\\reflections}] at (axis cs: 15.7, 92.3) {};
	
	\end{axis}
	\end{tikzpicture}
	
	\caption{Example of reflections at the underbody of a truck resulting in clutter. Coloring as in Fig.~\ref{fig:example_specular_reflection}.}
	\label{fig:example_underbody_reflections}
\end{figure}

In this work, clutter is defined as those moving detections in the output of a radar sensor whose position does not match the position of any real moving object (e.g. a car or pedestrian). A detection is said to be moving if the absolute value of its velocity after ego motion compensation $|v_\text{abs}|$ is above a certain threshold. The exclusion of stationary (i.e. nonmoving) detections is necessary since even if there isn't any object at their position, it is impossible to decide whether they result from one of the effects described in the following or stem from reflections at the ground. Micro-Doppler effects (see e.g.~\cite{Chen2006}) and ordinary measurement errors within the sensor's specification are tolerated, meaning that it's sufficient for detections to lie inside the respective margin around an object.

Many different effects can cause the occurrence of clutter. Examples include environmental noise, interference between multiple sensors, erroneous ambiguity resolution by the sensor's signal processing and multipath propagation. The latter can lead to a wide range of observations at the sensor due to arbitrary interference, waves traveling unexpected paths etc. Of these, three specific, well-defined effects commonly occurring in real traffic data shall be the focus in the following. The clutter resulting from them is especially hard to detect since it is often consistent over time and occurs in groups of multiple detections lying close to each other.

\subsection{Reflection at the Ego Vehicle}
The first effect of special interest is reflection at the ego vehicle. More specifically, a common occurrence for radar sensors mounted on a car is that the transmitted signal bounces multiple times between an object and the ego vehicle itself, before being received again. Every reflection at the ego vehicle means that the distance towards the object must be covered another time and any frequency shift introduced by the Doppler effect arises again. For the corresponding detection, the measured distance $d_\text{ref}$ and relative velocity $v_{\text{rel}, \text{ref}}$ (derived from the frequency shift) are thus integer multiples of the respective values measured for the direct path between sensor and object, i.e.
\begin{align}
d_\text{ref} &\approx (n + 1) \cdot d_\text{dir} \label{eq:d_reflection_at_ego}\\
v_{\text{rel}, \text{ref}} &\approx (n + 1) \cdot v_{\text{rel}, \text{dir}} \;, \label{eq:v_rel_reflection_at_ego}
\end{align}
where $n \in \mathds{N}$ is the number of times the signal was reflected at the ego vehicle. Fig.~\ref{fig:example_reflections_at_ego_vehicle} shows an example of the described effect.

\subsection{Underbody Reflections}
Another frequent observation are reflections between different parts of the underbody of a vehicle and/or the ground. This happens especially often at trucks. While theses reflections have negligible influence on the measured velocity, the distance of travel can be significantly increased. Thus, detections lying behind the reflecting vehicle can occur. An example of this is shown in Fig.~\ref{fig:example_underbody_reflections}.

\subsection{Multipath Propagation via Specular Reflection} \label{subsection:specular_reflection}
The last effect to be presented in detail is multipath propagation via specular reflection, often simply called multipath propagation. The term is used to describe an indirect propagation path between sensor and object via a reflecting surface. For example, the transmitted signal could first travel towards a guardrail positioned parallel to the road, be (partially) reflected and continue its way onto a vehicle, where it is reflected again and takes the same way back. The resulting detection would then lie in direction of the reflection point on the guardrail and have a distance greater than the vehicle's distance from the sensor. A real sensor measurement where the effect can be observed is depicted in Fig.~\ref{fig:example_specular_reflection}. There, the guardrails between lanes reflect the preceding vehicle, causing clutter on the opposite lane.

All surfaces that are smooth relative to the signal's wavelength cause specular reflections. The ratio of reflected energy depends on the surface's material. In automotive scenarios, multipath clutter typically occurs near concrete walls, guardrails, noise cancellation walls and similar objects.

Clutter resulting from specular reflection is categorized as in~\cite{Kraus2020}. Accordingly, a detection is said to be of type-1 or type-2 depending on whether the last reflection before the signal was received occurred at the target object or another surface. It is further denoted as $n$-bounce, where $n$ is the total number of times the signal was reflected. Thus, the example given above describes a type-2 3-bounce reflection.

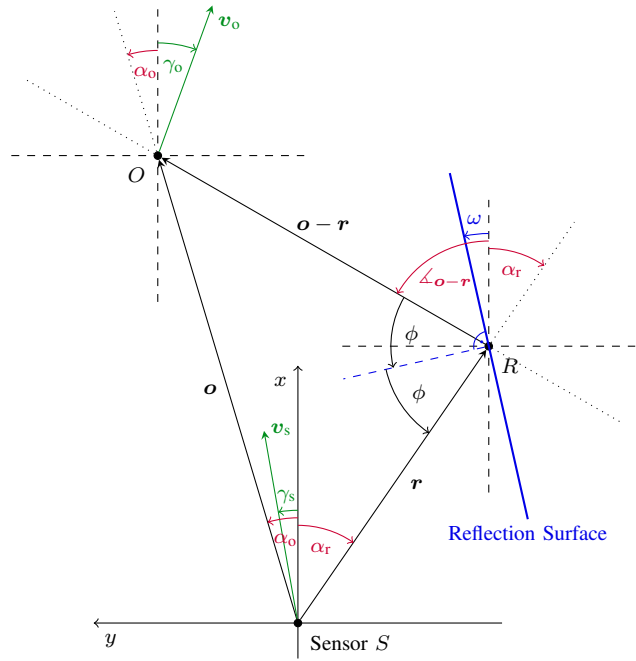
\begin{figure}[!t]
	\centering
	\begin{tikzpicture}
[
	scale = 1.18,
	every label/.append style = {font=\footnotesize}
]

\pgfmathsetmacro{\rd}{3.78987}
\pgfmathsetmacro{\razi}{55.43775}
\pgfmathsetmacro{\od}{5.50141}
\pgfmathsetmacro{\oazi}{106.6773}

\pgfmathsetmacro{\vs}{2.2}
\pgfmathsetmacro{\gammas}{100.0}
\pgfmathsetmacro{\vo}{1.8}
\pgfmathsetmacro{\gammao}{70.0}

\pgfmathsetmacro{\axisHelplineLength}{1.65}
\pgfmathsetmacro{\vectorHelplineLength}{1.7}

\pgfmathsetmacro{\reflectionSurfaceNormalLength}{1.68}
\pgfmathsetmacro{\halfReflectionSurfaceLength}{2.0}

\pgfmathsetmacro{\azimuthAngleRadius}{1.4}
\pgfmathsetmacro{\velocityAngleRadius}{1.5}
\pgfmathsetmacro{\reflectionSurfaceAngleRadius}{1.5}
\pgfmathsetmacro{\reflectionAngleRadius}{1.4}

\colorlet{azimuthColor}{purple!80!red}
\colorlet{velocityColor}{tpGreenColor}
\colorlet{reflectionSurfaceColor}{blue!90!black}

\tikzset{point/.style = {draw, circle, fill=black, inner sep=1pt}}
\tikzset{vector/.style = {draw, -stealth}}
\tikzset{vectorhelpline/.style = {draw, dotted, thin}}
\tikzset{axis/.style = {draw, ->}}
\tikzset{axishelpline/.style = {draw, dashed}}
\tikzset{textlabel/.style = {inner sep=2pt, font=\footnotesize, align=center}}

\node (s) at (0,0) [point, label={below right:Sensor $S$}]{};
\node (r) at +(\razi:\rd) [point, label={below right:$R$}]{};
\node (o) at +(\oazi:\od) [point, label={below left:$O$}]{};

\draw[axis] (s) ++(2.3, 0) -- ++(-4.6, 0) node[inner sep=0pt, label={below right:$y$}]{};
\draw[axis] (s) ++(0, -0.4) -- ++(0, 3.3) node[inner sep=0pt, label={below left:$x$}]{};
\draw[axishelpline] (r) ++(-\axisHelplineLength, 0) -- ++(2*\axisHelplineLength, 0);
\draw[axishelpline] (r) ++(0, -\axisHelplineLength) -- ++(0, 2*\axisHelplineLength);
\draw[axishelpline] (o) ++(-\axisHelplineLength, 0) -- ++(2*\axisHelplineLength, 0);
\draw[axishelpline] (o) ++(0, -\axisHelplineLength) -- ++(0, 2*\axisHelplineLength);

\draw[vector] (s) -- node[xshift=-0.02cm, label={right:$\boldsymbol{r}$}]{} (r);
\draw[vector] (s) -- node[xshift=0.1cm, label={left:$\boldsymbol{o}$}]{} (o);
\draw[vector] (r) -- node[label={above:$\boldsymbol{o} - \boldsymbol{r}$}]{} (o);

\draw[vectorhelpline] (r) -- ++(\razi:\vectorHelplineLength);
\draw[vectorhelpline] (o) -- ++(\oazi:\vectorHelplineLength);
\draw[vectorhelpline] ($ (r) ! -\vectorHelplineLength cm ! (o) $) -- (r);
\draw[vectorhelpline] ($ (o) ! -\vectorHelplineLength cm ! (r) $) -- (o);

\coordinate (aux1) at ($ (r) ! 1cm ! (o) $);
\coordinate (aux2) at ($ (r) ! 1cm ! (s) $);
\coordinate (auxRsNormal) at ($ (aux1) + (aux2) - (r) $);
\draw[dashed, reflectionSurfaceColor] (r) -- ($ (r) ! \reflectionSurfaceNormalLength cm ! (auxRsNormal) $);

\coordinate (auxRs1) at ($ (r) ! \halfReflectionSurfaceLength cm ! -90:(auxRsNormal) $);
\coordinate (auxRs2) at ($ (r) ! \halfReflectionSurfaceLength cm ! 90:(auxRsNormal) $);
\draw[thick, reflectionSurfaceColor] (auxRs1) -- (auxRs2) node[textlabel, xshift=-0.0cm, below]{Reflection Surface};

\coordinate (auxVelocityS) at ($ (s) + (\gammas:\vs) $);
\draw[vector, color=velocityColor] (s) -- (auxVelocityS) node[inner sep=0pt, xshift=-0.06cm, yshift=0.02cm, label={right:$\boldsymbol{v}_\text{s}$}]{};
\coordinate (auxVelocityO) at ($ (o) + (\gammao:\vo) $);
\draw[vector, color=velocityColor] (o) -- (auxVelocityO) node[inner sep=0pt, xshift=-0.07cm, label={below right:$\boldsymbol{v}_\text{o}$}]{};

\draw pic[draw=reflectionSurfaceColor, angle eccentricity=0.55, angle radius=0.2cm, pic text={.}, pic text options={color=reflectionSurfaceColor, inner sep=0pt, align=center}] {angle=auxRs1--r--auxRsNormal};

\coordinate (aux1) at ($ (r) + (0, 1) $);
\draw pic[draw=reflectionSurfaceColor, ->, angle eccentricity=1.13, angle radius=\reflectionSurfaceAngleRadius cm, pic text={$\omega$}, pic text options={color=reflectionSurfaceColor, font=\footnotesize}] {angle=aux1--r--auxRs1};

\draw pic[draw, ->, angle eccentricity=0.8, angle radius=\reflectionAngleRadius cm - 0.1cm, pic text={$\phi$}, pic text options={font=\footnotesize}] {angle=o--r--auxRsNormal};
\draw pic[draw, ->, angle eccentricity=0.8, angle radius=\reflectionAngleRadius cm, pic text={$\phi$}, pic text options={font=\footnotesize}] {angle=auxRsNormal--r--s};

\coordinate (aux1) at ($ (s) + (0, 1) $);
\draw pic[draw=azimuthColor, ->, angle eccentricity=0.8, angle radius=\azimuthAngleRadius cm, pic text={$\alpha_\text{o}$}, pic text options={color=azimuthColor, font=\footnotesize}] {angle=aux1--s--o};
\draw pic[draw=azimuthColor, <-, angle eccentricity=0.8, angle radius=\azimuthAngleRadius cm - 0.1cm, pic text={$\alpha_\text{r}$}, pic text options={color=azimuthColor, font=\footnotesize}] {angle=r--s--aux1};

\coordinate (aux1) at ($ (o) + (0, 1) $);
\coordinate (aux2) at ($ (o) + (\oazi:1) $);
\draw pic[draw=azimuthColor, ->, angle eccentricity=0.8, angle radius=\azimuthAngleRadius cm, pic text={$\alpha_\text{o}$}, pic text options={color=azimuthColor, font=\footnotesize}] {angle=aux1--o--aux2};

\coordinate (aux1) at ($ (r) + (0, 1) $);
\coordinate (aux2) at ($ (r) + (\razi:1) $);
\draw pic[draw=azimuthColor, <-, angle eccentricity=0.8, angle radius=\azimuthAngleRadius cm - 0.1cm, pic text={$\alpha_\text{r}$}, pic text options={color=azimuthColor, font=\footnotesize}] {angle=aux2--r--aux1};
\draw pic[draw=azimuthColor, ->, angle eccentricity=0.75, angle radius=\azimuthAngleRadius cm, pic text={$\measuredangle_{\boldsymbol{o} - \boldsymbol{r}}$}, pic text options={color=azimuthColor, font=\footnotesize, xshift=-0.05cm}] {angle=aux1--r--o};

\coordinate (aux1) at ($ (s) + (0, 1) $);
\draw pic[draw=velocityColor, ->, angle eccentricity=1.13, angle radius=\velocityAngleRadius cm, pic text={$\gamma_\text{s}$}, pic text options={color=velocityColor, font=\footnotesize}] {angle=aux1--s--auxVelocityS};

\coordinate (aux1) at ($ (o) + (0, 1) $);
\draw pic[draw=velocityColor, <-, angle eccentricity=0.8, angle radius=\velocityAngleRadius cm, pic text={$\gamma_\text{o}$}, pic text options={color=velocityColor, font=\footnotesize}] {angle=auxVelocityO--o--aux1};

\node[draw=white, fill=white, inner sep=0pt, outer sep=0pt, above=0.1cm of auxVelocityO] {};

\end{tikzpicture}
	\caption{Geometric model of multipath propagation via specular reflection. $\boldsymbol{o}$ and $\boldsymbol{r}$ are the vectors from sensor $S$ to reflection points $O$ on the target object and $R$ on the reflection surface, $\omega \in \left[-\frac{\pi}{2}, \frac{\pi}{2}\right)$ is the surface's orientation and $\phi \in \left[-\pi, \pi\right)$ the reflection angle. Sensor $S$ and reflection point $O$ move with velocities $\boldsymbol{v}_\text{s}$ and $\boldsymbol{v}_\text{o}$, respectively.}
	\label{fig:specular_reflection_geometry}
\end{figure}
If the position and velocity of all involved objects are known, the respective values theoretically measured for a clutter detection caused by specular reflection can be calculated. Using the variables introduced in Fig.~\ref{fig:specular_reflection_geometry}, the desired direct path (i.e. type-1 1-bounce) detection corresponding to the target object has distance, azimuth angle, relative radial velocity and ego motion compensated absolute radial velocity
\begin{align}
d_{11} &= \norm{\boldsymbol{o}}_2 = o\\
\alpha_{11} &= \alpha_\text{o}\\
v_{\text{rel}, 11} &= v_\text{o} \cdot \cos(\gamma_\text{o} - \alpha_\text{o}) - v_\text{s} \cdot \cos(\gamma_\text{s} - \alpha_\text{o})\\
v_{\text{abs}, 11} &= v_\text{o} \cdot \cos(\gamma_\text{o} - \alpha_\text{o}) \label{eq:v_abs_11} \;,
\end{align}
where $x = \norm{\boldsymbol{x}}_2$ is the length of a vector $\boldsymbol{x}$. In contrast, a type-2 3-bounce reflection leads to a detection with
\begin{align}
d_{23} &= r + \norm{\boldsymbol{o} - \boldsymbol{r}}_2 \label{eq:d_23}\\
\alpha_{23} &= \alpha_\text{r}\\
v_{\text{rel}, 23} &= v_\text{o} \cdot \cos(\gamma_\text{o} - \measuredangle_{\boldsymbol{o} - \boldsymbol{r}}) - v_\text{s} \cdot \cos(\gamma_\text{s} - \alpha_\text{r})\\
v_{\text{abs}, 23} &= v_\text{o} \cdot \cos(\gamma_\text{o} - \measuredangle_{\boldsymbol{o} - \boldsymbol{r}}) \label{eq:v_abs_23} \;,
\end{align}
where $\measuredangle_{\boldsymbol{o} - \boldsymbol{r}} = \pi + \alpha_\text{r} - 2\phi = 2\omega - \alpha_\text{r}$. The reason that these values are measured is that the transmitted signal has to travel more distance and arrives at the receiver coming from the reflection surface. Note that the relative velocity must be calculated regarding the components of $\boldsymbol{v}_\text{s}$ and $\boldsymbol{v}_\text{o}$ that point in the direction of propagation. For type-1 and -2 2-bounce reflections, the measured values accordingly are
\begin{align}
d_{12} = d_{22} &= \tfrac{1}{2} \left( o + r + \norm{\boldsymbol{o} - \boldsymbol{r}}_2 \right) \label{eq:d_X2}\\
\alpha_{12} &= \alpha_\text{o} \label{eq:alpha_12}\\
\alpha_{22} &= \alpha_\text{r} \label{eq:alpha_22}\\
v_{\text{rel}, 12} = v_{\text{rel}, 22} &= \tfrac{1}{2} v_{\text{rel}, 11} + \tfrac{1}{2} v_{\text{rel}, 23} \label{eq:v_rel_X2} \;.
\end{align}
Here, the ego motion component in \eqref{eq:v_rel_X2} cannot be successfully eliminated during compensation, since only the azimuth angle that was actually measured is known to the respective module. The determined absolute radial velocities are therefore
\begin{align}
\begin{split}
	\label{eq:v_abs_12}
	v_{\text{abs}, 12} &= v_{\text{rel}, 12} + v_\text{s} \cdot \cos(\gamma_\text{s} - \alpha_\text{o})\\
	&= \tfrac{1}{2} v_\text{o} \left(\cos(\gamma_\text{o} - \alpha_\text{o}) + \cos(\gamma_\text{o} - \measuredangle_{\boldsymbol{o} - \boldsymbol{r}})\right)\\
	&\qquad + \tfrac{1}{2} v_\text{s} \left(\cos(\gamma_\text{s} - \alpha_\text{o}) - \cos(\gamma_\text{s} - \alpha_\text{r})\right)
\end{split}
\\
\begin{split}
	\label{eq:v_abs_22}
	v_{\text{abs}, 22} &= v_{\text{rel}, 22} + v_\text{s} \cdot \cos(\gamma_\text{s} - \alpha_\text{r})\\
	&= \tfrac{1}{2} v_\text{o} \left(\cos(\gamma_\text{o} - \alpha_\text{o}) + \cos(\gamma_\text{o} - \measuredangle_{\boldsymbol{o} - \boldsymbol{r}})\right)\\
	&\qquad + \tfrac{1}{2} v_\text{s} \left(\cos(\gamma_\text{s} - \alpha_\text{r}) - \cos(\gamma_\text{s} - \alpha_\text{o})\right) \;.
\end{split}
\end{align}
The described relations for measured distances and azimuth angles are experimentally validated in~\cite{Kamann2018}. It is also shown that these three kinds of clutter can occur in any combination in practice. The formulas hold for all constellations (e.g. reflection surface on the left of the sensor) and can easily be extended to include moving reflection points $R$. Four- and higher-bounce clutter isn't discussed since it cannot be observed in the data recorded for this paper.

\section{Algorithm for Identifying Clutter Detections} \label{section:methods}

\begin{figure}[!t]
	\centering
	\begin{tikzpicture}
[
	scale = 1.0,
	every label/.append style = {font=\footnotesize}
]

\pgfmathsetmacro{\horizontalD}{0.8}
\pgfmathsetmacro{\verticalD}{0.4}
\pgfmathsetmacro{\nodeWidth}{3.2}
\pgfmathsetmacro{\bufferDiameter}{1.2}

\definecolor{inoutColor}{HTML}{FFD5D5}
\definecolor{checkColor}{HTML}{AFDDE9}
\definecolor{bufferColor}{HTML}{DECD87}

\tikzset{inout/.style = {draw, ellipse, fill=inoutColor, minimum width=\nodeWidth cm, minimum height=0.75cm, inner sep=1pt, align=center, font=\footnotesize}}
\tikzset{check/.style = {draw, rectangle, rounded corners=0.25cm, fill=checkColor, minimum width=\nodeWidth cm, minimum height=0.75cm, inner sep=1pt, align=center, font=\footnotesize}}
\tikzset{line/.style = {draw, -stealth}}

\node[inout] (inDetections) {detections};
\node[inout, right=\horizontalD cm of inDetections] (inEgoMotion) {ego motion data};

\node[check, below=\verticalD cm of inDetections] (filtering) {filter by\\individual values};
\node[check, below=\verticalD cm of inEgoMotion] (search) {search for\\similar detections};
\path[line] (inDetections) -- (filtering);
\path[line] (inEgoMotion) -- (search);
\path[line] (filtering) -- (search);

\node[check, below=\verticalD cm of search] (refAtEgo) {check for reflections\\at ego vehicle};
\path[line] (search) -- (refAtEgo);

\node[circle, draw=black, fill=bufferColor, left=\horizontalD cm of refAtEgo, minimum size=\bufferDiameter cm] (outerCircle) {};
\foreach \angle in {45, 135}
	\draw[black] (outerCircle) +(\angle:\bufferDiameter/2) -- +(\angle - 180:\bufferDiameter/2);
\node[circle, draw=black, fill=white, align=center, font=\scriptsize, inner sep=0.5pt, minimum size=\bufferDiameter cm - 0.3cm] (innerCircle) at (outerCircle.center) {ring\\buffer};
\path[line, shorten <= -0.05mm] (filtering) -- (outerCircle);
\path[line, shorten >=-1.11mm] (outerCircle) -- (search.south west);

\node[check, below=\verticalD cm of refAtEgo] (underbodyRef) {check for\\underbody reflections};
\node[inout, left=\horizontalD cm of underbodyRef] (inSurfaces) {positions of\\reflective surfaces};
\path[line] (refAtEgo) -- (underbodyRef);

\node[check, below=\verticalD cm of underbodyRef] (specularRef) {check for\\specular reflections};
\node[inout, left=\horizontalD cm of specularRef] (out) {classification};
\path[line] (underbodyRef) -- (specularRef);
\path[line, shorten <= -0.05mm, shorten >=-1.21mm] (inSurfaces.south east) -- (specularRef.north west);
\path[line] (specularRef) -- (out);

\node[draw=white, fill=white, inner sep=0pt, outer sep=0pt, above=0.1cm of inDetections] {};
\node[draw=white, fill=white, inner sep=0pt, outer sep=0pt, below=0.1cm of out] {};

\end{tikzpicture}
	\caption{Overview of the developed algorithm}
	\label{fig:algorithm_overview}
\end{figure}
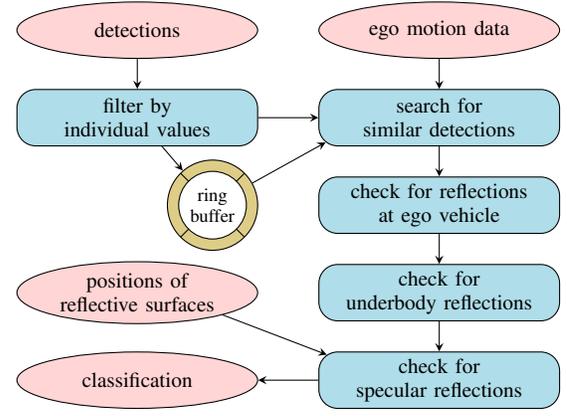

The developed algorithm is mainly based on the description of propagation paths that lead to clutter in the previous section. It determines for each moving detection in a sensor measurement whether or not it is clutter. A naive filtering based on the values of each individual detection and a search for similar other detections in the current and previous measurements serve as a way to detect easy-to-find, unsystematic clutter. Three checks specifically tailored towards the described clutter effects follow. Each is aimed at identifying detections resulting from one of them by analyzing whether the respective corresponding propagation path is likely to have occurred. If a check classifies a detection as clutter, all subsequent checks for this detection are skipped. Hence, the checks are ordered by their computational cost and likelihood of succeeding. An overview of the algorithm is given in Fig.~\ref{fig:algorithm_overview}.

The main input to the algorithm are the detections generated from a single sensor measurement. Each detection is a tuple of the measured values for distance $d$, azimuth angle $\alpha$, relative radial velocity $v_\text{rel}$, ego-motion-compensated absolute radial velocity $v_\text{abs}$ and radar cross-section $\mathit{rcs}$ of the reflecting object. In addition to that, information about the ego vehicle's movement and a list of positions of potentially reflecting surfaces like guardrails or concrete walls are required. The latter can be generated in a preceding processing step either purely from radar data or e.g. using lidar as in~\cite{Scheiner2020}. In this work, a method based on clustering stationary radar detections and searching for line segments inside clusters is utilized. Clustering uses the algorithm presented in~\cite{RBNN}, line segments are found with an adapted RANSAC method.

\subsection{Filtering and Search for Similar Detections}
The first step in the algorithm is a naive filtering performed individually for each detection. Depending on the measured distance $d$, a threshold for the value of $\mathit{rcs}$ is set, below which the respective detection is excluded from all further processing. Affected moving detections are additionally classified as clutter.

Next, for every moving detection, the current and a small number of buffered previous measurements are searched for detections with similar position and velocity. The detection under test (DUT) is assumed to be clutter unless a certain number of such detections is found. Since road users can move between measurements, directly comparing the DUT to older detections is inaccurate. Instead, the position which the reflection point that corresponds to the DUT theoretically had at the moment the earlier measurement was taken is used for comparison. It is calculated from the DUT's absolute radial velocity, the available ego motion data and the time elapsed. The area around this position, in which other detections must lie, is dynamically adjusted in size according to the examined situation. In particular, the time elapsed between the compared measurements and an estimate of the reflection point's maximum tangential movement are taken into account.

\subsection{Identification of Clutter Caused by Reflection at the Ego Vehicle}
To identify clutter resulting from reflection at the ego vehicle, the DUT is compared to other moving detections in the same measurement. Should one with similar azimuth $\alpha_\text{dir} \approx \alpha_\text{dut}$ for which the equations \eqref{eq:d_reflection_at_ego} and \eqref{eq:v_rel_reflection_at_ego} hold exist, the DUT is classified as clutter. Even though reflections between stationary objects and the ego vehicle can also lead to clutter, only moving detections are considered. The aim of this restriction is to prevent false positives when a stationary object or vegetation is located in the middle between sensor and opposite lane. To reduce the amount of errors even further, $n$ is limited to an expected maximum number of reflections. A relaxation is applied in case both $v_{\text{rel}, \text{dut}} \approx 0$ and $v_{\text{rel}, \text{dir}} \approx 0$. Then, only the conditions for azimuth and distance have to be met, reducing the influence of measurement errors when the ego vehicle and another road user move at roughly the same speed.

\subsection{Identification of Clutter Caused by Underbody Reflections}
Next, it is tested whether the DUT is likely to be clutter resulting from reflections at the underbody of a vehicle. Since no clearly defined relations between detections could be formulated here, a heuristic method is employed. To this end, detections with similar azimuth and velocity as the DUT are identified. If the number of matching detections that lie slightly closer to the sensor is above a certain threshold and the number of matches with slightly higher distance below another one, the DUT is classified as clutter. This criterion is motivated by the observation that clutter caused by underbody reflections is typically located behind multiple nonclutter detections that correspond to the reflecting vehicle.

\subsection{Identification of Clutter Caused by Specular Reflection}
The method for detecting clutter that results from multipath propagation via specular reflection is executed last. It is based on the assumption that, should the DUT be such clutter, it is possible to find a point $O$ on an object and a point $R$ on a reflective surface for which the relations described in Section~\ref{subsection:specular_reflection} hold. Thus, a search for these points is conducted. For lack of more accurate information about the location of objects, the only points that come into consideration for $O$ are the positions of detections in the current measurement. If fitting points $O$ and $R$ are found, the respective propagation path is considered to be sufficiently likely and the DUT is classified as clutter.

Candidates for the two points are determined as follows. When testing whether the DUT is clutter stemming from a type-1 reflection, candidate $O'$ must be a detection lying between sensor $S$ and DUT, i.e. $\alpha_{\text{o}'} \approx \alpha_\text{dut}$, $d_{\text{o}'} < d_\text{dut}$. After choosing $O'$, fitting candidates $R'$ are found by searching for points on the known surfaces for which the angle of incidence is equal to the angle of reflection. This corresponds to solving
\begin{align}
\phi_1 = \frac{\pi}{2} + \alpha_{\text{r}'} - \omega' \overset{!}{=} \frac{\pi}{2} - \measuredangle_{\boldsymbol{o}' - \boldsymbol{r}'} + \omega' = \phi_2 \;, \label{eq:reflection_angle_comparison}
\end{align}
where $\omega'$ is the currently examined surface's orientation. When searching for type-2 clutter, $R'$ must be the intersection of the vector pointing from $S$ to DUT and one of the surfaces. $O'$ can then be any detection for which the left- and right-hand side of \eqref{eq:reflection_angle_comparison} are approximately equal. In both cases stationary detections are included when looking for $O'$. This is because objects moving tangentially to the sensor's line of sight appear stationary if observed via the direct propagation path but moving if seen via specular reflection. Moreover, 2-bounce reflections of stationary objects can seem to be in motion due to the erroneous ego motion compensation in \eqref{eq:v_abs_12} and \eqref{eq:v_abs_22}.

Once candidates $O'$ and $R'$ have been chosen, the distance that would be measured for a clutter detection resulting from this propagation path is calculated via \eqref{eq:d_23} or \eqref{eq:d_X2}, depending on the investigated case. The determined value must be close to $d_\text{dut}$.
Since the movement direction $\gamma_{\text{o}'}$ of the object presumably represented by $O'$ isn't known, checking the velocity requires more steps. Especially in extra-urban traffic it is justified to assume a maximum possible difference in direction of movement $\delta_\text{max} \in [0, \tfrac{\pi}{2}]$ between the ego vehicle and other road users. This limits $\gamma_{\text{o}'}$ to values in the set
\begin{align}
\begin{split}
\Gamma_1 &= \left[ -\psi_\text{s} - \delta_\text{max}, -\psi_\text{s} + \delta_\text{max} \right]\\
&\qquad \cup \left[ \pi - \psi_\text{s} - \delta_\text{max}, \pi - \psi_\text{s} + \delta_\text{max} \right] \;,
\end{split}
\end{align}
where $\psi_\text{s}$ is the radar sensor's yaw relative to the ego vehicle. To restrict $\gamma_{\text{o}'}$ further, a maximum realistic velocity $v_\text{max}$ of objects is assumed. Applying $v_{\text{o}'} \leq v_\text{max}$ to \eqref{eq:v_abs_11} leads to
\begin{align}
\begin{split}
	\Gamma_2 &= \left[ \alpha_{\text{o}'} - \arccos \left( \frac{|v_{\text{abs}, \text{o}'}|}{v_\text{max}} \right) + \pi \cdot \mathrm{u}(-v_{\text{abs}, \text{o}'}), \right.\\
	&\qquad \left. \alpha_{\text{o}'} + \arccos \left( \frac{|v_{\text{abs}, \text{o}'}|}{v_\text{max}} \right) + \pi \cdot \mathrm{u}(-v_{\text{abs}, \text{o}'}) \right] \;,
\end{split}
\end{align}
where $\mathrm{u}(.)$ is the unit step function. The case $|v_{\text{abs}, \text{o}'}| > v_\text{max}$ is handled by setting $\Gamma_2 = \{\alpha_{\text{o}'} + \pi \cdot \mathrm{u}(-v_{\text{abs}, \text{o}'})\}$. Stationary detections, i.e. ones with very low $|v_{\text{abs}, \text{o}'}|$, must be handled separately to cover both tangential movement and objects actually standing still.
By combining \eqref{eq:v_abs_11} and \eqref{eq:v_abs_23},
\begin{align}
v'_{\text{abs}, 23} = v_{\text{abs}, \text{o}'} \cdot \frac{\cos(\gamma_{\text{o}'} - \measuredangle_{\boldsymbol{o}' - \boldsymbol{r}'})}{\cos(\gamma_{\text{o}'} - \alpha_{\text{o}'})} \label{eq:v_abs_23_dash}
\end{align}
is derived.
This expression is monotonic w.r.t. $\gamma_{\text{o}'} \in \Gamma_1 \cap \Gamma_2$. Thus, substituting the limits determined for $\gamma_{\text{o}'}$ into \eqref{eq:v_abs_23_dash} directly results in limits for an interval $V$. $V$ contains all values the velocity component $v'_{\text{abs}, 23}$ of $v_{\text{o}'}$ in direction $\measuredangle_{\boldsymbol{o}' - \boldsymbol{r}'}$ could have, given the mentioned assumptions. Only if the absolute velocity measured for the DUT $v_{\text{abs}, \text{dut}}$ is included in $V$, it is possible that the DUT is type-2 3-bounce clutter resulting from the respective propagation path over $O'$ and $R'$. Checking for 2-bounce clutter is done similarly, only that the interval limits must be scaled by $\tfrac{1}{2}$ and shifted by an offset calculated from the involved detections' velocities.

\section{Evaluation} \label{section:evaluation}

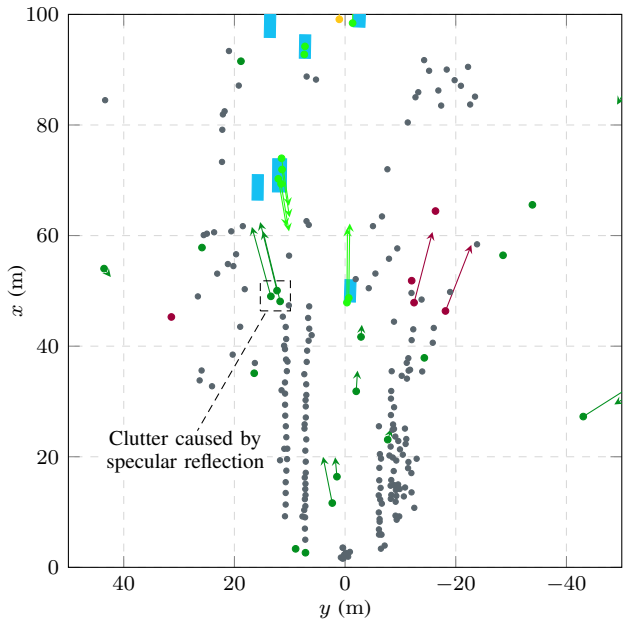
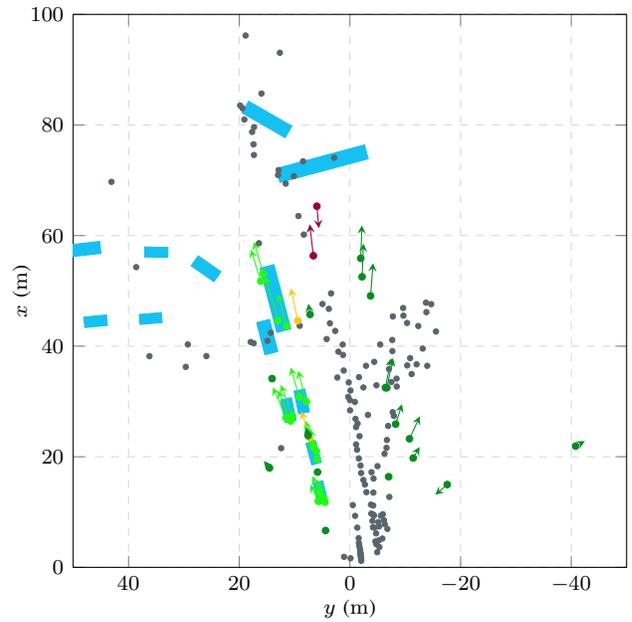
\begin{figure*}[!t]
	\begin{subfigure}{\columnwidth}
		\hspace*{-0.1cm}
		\begin{tikzpicture}
		[
		every axis/.append style = {label style={font=\footnotesize}, tick label style={font=\footnotesize}},
		every pin/.style = {font=\footnotesize, align=center},
		every pin edge/.style = {black, densely dashed}
		]
		\begin{axis}
		[
		width=1.20*\columnwidth,
		grid=major,
		grid style={dashed,gray!30},
		xlabel= $y$ (\si{\meter}),
		ylabel= $x$ (\si{\meter}),
		x label style={yshift=0.15cm},
		x dir=reverse,
		y label style={yshift=-0.25cm},
		axis equal image,
		xmin=-50,
		xmax=50,
		ymax=100,
		ymin=0,
		colormap name=cm_radar_dl_confusion
		]

		\drawObject{52.200793227351255}{-0.9938576201745946}{-1.8798025420001778}{-0.2}{-4.2}{1.0}{-1.0}
		\drawObject{68.71518440817454}{15.791115263230267}{179.2757036388099}{2.3}{-2.3}{1.0}{-1.0}
		\drawObject{69.88644591767513}{11.855546439145883}{179.49469017219985}{2.0}{-4.0}{1.25}{-1.25}
		\drawObject{94.18646582261442}{7.23804598068773}{179.0}{2.12}{-2.12}{1.0}{-1.0}
		\drawObject{97.95252970821866}{13.567980927024792}{178.8976706658552}{2.12}{-2.12}{1.0}{-1.0}
		\drawObject{99.83184245131838}{-2.6}{-2.1488226140260283}{2.12}{-2.12}{1.0625}{-1.0625}
		
		\plotRadarDetectionListConfusion{resources/qualitative_evaluation_positive_example_specular_reflection_dl.csv}
		
		\node[rectangle, draw=black, densely dashed, minimum width=0.4cm, minimum height=0.4cm, inner sep=0pt, outer sep=1pt, font=\footnotesize, pin={[fill=white, inner sep=1pt, outer sep=0pt, anchor=center, pin distance=2.12cm, pin edge={shorten >= -0.05cm}] -120.0:Clutter caused by\\specular reflection}] at (axis cs: 12.6, 49.1) {};
		
		\end{axis}
		\end{tikzpicture}
		
		\caption{Output for the situation shown in Fig.~\ref{fig:example_specular_reflection}. Most detections, including the ones resulting from specular reflections at the guardrails between lanes, are correctly classified.}
		\label{fig:qualitative_evaluation_positive_example_1}
	\end{subfigure}
	\hfill
	\begin{subfigure}{\columnwidth}
		\hspace*{-0.1cm}
		\begin{tikzpicture}
		[
		every axis/.append style = {label style={font=\footnotesize}, tick label style={font=\footnotesize}},
		every pin/.style = {font=\footnotesize, align=center},
		every pin edge/.style = {black, densely dashed}
		]
		\begin{axis}
		[
		width=1.20*\columnwidth,
		grid=major,
		grid style={dashed,gray!30},
		xlabel= $y$ (\si{\meter}),
		ylabel= $x$ (\si{\meter}),
		x label style={yshift=0.15cm},
		x dir=reverse,
		y label style={yshift=-0.25cm},
		axis equal image,
		xmin=-50,
		xmax=50,
		ymax=100,
		ymin=0,
		colormap name=cm_radar_dl_confusion
		]
		
		\drawObject{13.505388742378727}{5.318231796873505}{13.191741198697823}{2.0}{-2.0}{0.9}{-0.9}
		\drawObject{20.68561703284763}{6.546684421716037}{15.0}{2.0}{-2.0}{0.9}{-0.9}
		\drawObject{28.422355798620288}{11.12735911274902}{12.0}{2.1}{-2.1}{1.0}{-1.0}
		\drawObject{30.0}{8.7}{11.5}{2.1}{-2.1}{1.0}{-1.0}
		\drawObject{42.6054415023973}{15.187042016274319}{14.398172651770874}{2.0}{-4.0}{1.2}{-1.2}
		\drawObject{46.855595656867195}{12.689355768215705}{14.727758282142682}{7.9}{-4.1}{1.6}{-1.0}
		\drawObject{54.0}{26.0}{55.0}{2.5}{-2.5}{1.1}{-1.1}
		\drawObject{57.0}{35.0}{89.0}{2.1}{-2.1}{0.9}{-0.9}
		\drawObject{57.54116789950649}{47.897209962249434}{96.53173985242229}{2.8125}{-2.8125}{1.087499976158142}{-1.087499976158142}
		\drawObject{45.0}{36.0}{-85.0}{2.0}{-2.0}{0.9}{-0.9}
		\drawObject{44.5}{46.0}{-84.5}{2.1}{-2.0}{0.95}{-0.95}
		\drawObject{73.0}{5.0}{105.0}{8.25}{-8.25}{1.3}{-1.3}
		\drawObject{81.0}{15.0}{60.0}{4.5}{-4.5}{1.2}{-1.2}
		
		\plotRadarDetectionListConfusion{resources/qualitative_evaluation_positive_example_waiting_in_line_dl.csv}
		
		\end{axis}
		\end{tikzpicture}
		
		\caption{A group of vehicles slowly approaching an intersection. Specular reflections occur at the bridge railing located on the right.\\
		\phantom{Invisible text for vertical alignment of subfigures.}}
		\label{fig:qualitative_evaluation_positive_example_2}
	\end{subfigure}
	
	\caption{Examples for qualitative evaluation. The color of a detection indicates the algorithm's output compared to the ground truth. Light and dark green detections are correctly predicted to be nonclutter and clutter, respectively. Light red detections correspond to real moving objects but are mistakenly classified as clutter, whereas dark red ones are clutter that wasn't detected. Gray and yellow points mark stationary detections and detections for which classification is ambiguous.}
	\label{fig:qualitative_evaluation_positive_examples}
\end{figure*}

\subsection{Data Set}

The data used for fine-tuning and evaluation of the designed algorithm was recorded with a mass-produced 2D \SI{77}{GHz} FMCW mid-range radar sensor that is commonly employed in commercially available cars. The sensor has about \SI{110}{m} range and $\pm \SI{60}{\SIUnitSymbolDegree}$ horizontal FoV. It was mounted on the front center of the test vehicle.
Recordings were made in real extra-urban traffic. 2,390 radar measurements from multiple sequences of consecutive time steps, containing a total of 585,434 detections, were later labeled. During annotation, it was distinguished between detections likely to stem from a stationary reflection point, detections that correspond to a moving object and clutter as defined in Section~\ref{section:clutter_effects}. Detections for which it is impossible to decide for one of these classes were put into a separate category. The final distribution of classes among all moving detections is given in Tab.~\ref{tab:class_distribution}.

\begin{table}[!b]
	\centering
	\caption{Class distribution of moving detections in data set}
	\label{tab:class_distribution}
	\begin{tabular}{ c | c | c  }
		Clutter & No Clutter & Ambiguous\\
		\hline
		\rule{0pt}{8pt} \SI{90.12}{\percent} & \SI{7.99}{\percent} & \SI{1.89}{\percent}
	\end{tabular}
\end{table}

Labeling was done in a semi-automatic way. To this end, an automated prelabeling method first classifies all detections with low absolute velocity as stationary. The remaining detections are then compared to the positions of objects detected by a tracking algorithm~\cite{Steyer2020}. Said algorithm combines both radar and lidar information and is executed in real-time during data recording. Any detection located directly inside an object's bounding box is labeled as nonclutter. Detections within the angle-dependent tolerances for measurement errors around an object are preliminarily classified as ambiguous. The employed tracking algorithm often underestimates the extend of objects in the direction facing away from the sensors. Therefore, detections on the unobservable side of an object which lie slightly outside of the tolerances are marked, indicating that they must be labeled manually. Finally, all moving detections not close to any tracked object are classified as clutter.

Following the described method for automatic prelabeling, annotations were manually corrected. In particular, it was checked whether the tracking algorithm missed or mistakenly added any objects and that it correctly determined their sizes. Marked detections and detections preliminarily labeled as ambiguous were assigned to the best fitting class.

\subsection{Results}

\newcommand\TP{\mathit{TP}}
\newcommand\FP{\mathit{FP}}
\newcommand\TN{\mathit{TN}}
\newcommand\FN{\mathit{FN}}

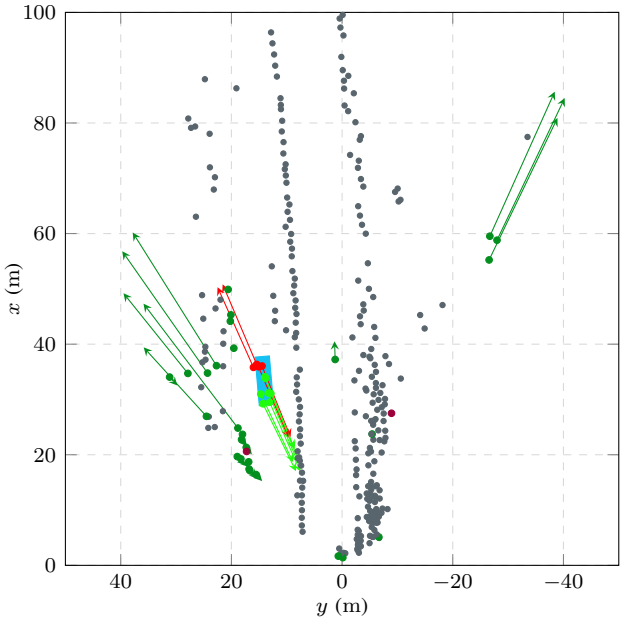
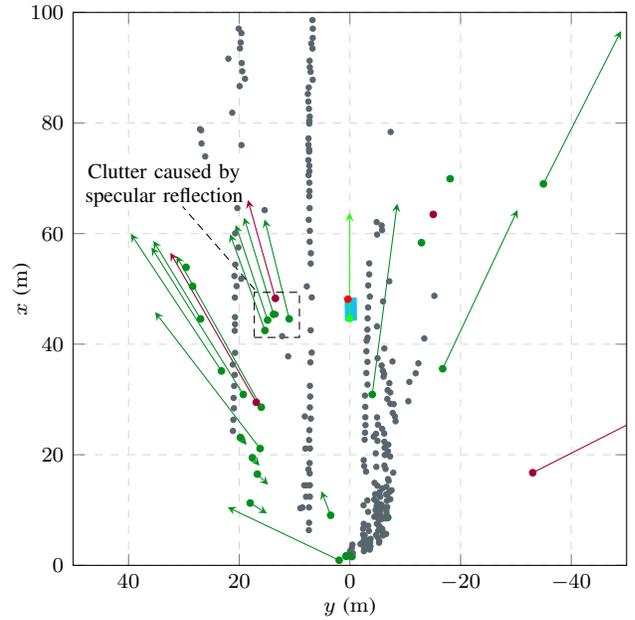
\begin{figure*}[!t]
	\begin{subfigure}{\columnwidth}
		\hspace*{-0.1cm}
		\begin{tikzpicture}
		[
			every axis/.append style = {label style={font=\footnotesize}, tick label style={font=\footnotesize}},
			every pin/.style = {font=\footnotesize, align=center},
			every pin edge/.style = {black, densely dashed}
		]
		\begin{axis}
		[
			width=1.20*\columnwidth,
			grid=major,
			grid style={dashed,gray!30},
			xlabel= $y$ (\si{\meter}),
			ylabel= $x$ (\si{\meter}),
			x label style={yshift=0.15cm},
			x dir=reverse,
			y label style={yshift=-0.25cm},
			axis equal image,
			xmin=-50,
			xmax=50,
			ymax=100,
			ymin=0,
			colormap name=cm_radar_dl_confusion
		]
		
			\drawObject{32.78493775361608}{14.0}{-175.0}{3.8125}{-5.0}{1.3}{-1.3}
			
			\plotRadarDetectionListConfusion{resources/qualitative_evaluation_negative_example_underbody_reflections_dl.csv}
		
		\end{axis}
		\end{tikzpicture}
		
		\caption{A truck on the opposite lane behind a guardrail. Several of the detections that correspond to the truck are mistakenly classified as clutter.}
		\label{fig:qualitative_evaluation_negative_example_1}
	\end{subfigure}
	\hfill
	\begin{subfigure}{\columnwidth}
		\hspace*{-0.1cm}
		\begin{tikzpicture}
		[
			every axis/.append style = {label style={font=\footnotesize}, tick label style={font=\footnotesize}},
			every pin/.style = {font=\footnotesize, align=center},
			every pin edge/.style = {black, densely dashed}
		]
		\begin{axis}
		[
			width=1.20*\columnwidth,
			grid=major,
			grid style={dashed,gray!30},
			xlabel= $y$ (\si{\meter}),
			ylabel= $x$ (\si{\meter}),
			x label style={yshift=0.15cm},
			x dir=reverse,
			y label style={yshift=-0.25cm},
			axis equal image,
			xmin=-50,
			xmax=50,
			ymax=100,
			ymin=0,
			colormap name=cm_radar_dl_confusion
		]
		
			\drawObject{46.321658653051486}{-0.16389747957100553}{0.6184692185018821}{2.0}{-2.0}{1.0}{-1.0}
			\drawObject{122.21349487917689}{-4.032724712836398}{-1.5510374847279162}{2.1624999046325684}{-2.1624999046325684}{1.2625000476837158}{-1.2625000476837158}
			\drawObject{-8.687235032028866}{13.677841552851532}{-179.94289832973757}{2.4124999046325684}{-2.4124999046325684}{1.087499976158142}{-1.087499976158142}
			
			\plotRadarDetectionListConfusion{resources/qualitative_evaluation_negative_example_underbody_vs_specular_dl.csv}
			
			\node[rectangle, draw=black, densely dashed, minimum width=0.6cm, minimum height=0.6cm, inner sep=0pt, outer sep=1pt, font=\footnotesize, pin={[fill=white, inner sep=1pt, outer sep=0pt, anchor=center, pin distance=1.84cm, pin edge={shorten >= -0.05cm}] 131.5:Clutter caused by\\specular reflection}] at (axis cs: 13.2, 45.3) {};
		
		\end{axis}
		\end{tikzpicture}
		
		\caption{A car on the same lane as the ego vehicle. Specular reflections occur at the guardrail between lanes.\\
		\phantom{Invisible text for vertical alignment of subfigures.}}
		\label{fig:qualitative_evaluation_negative_example_2}
	\end{subfigure}

	\caption{Examples for qualitative evaluation that highlight shortcomings of the proposed algorithm. Coloring as in Fig.~\ref{fig:qualitative_evaluation_positive_examples}.}
	\label{fig:qualitative_evaluation_negative_examples}
\end{figure*}

The presented algorithm for identifying clutter detections was evaluated using the described data set. During evaluation, the velocity threshold above which a detection is referred to as "moving" was empirically set to $|v_\text{abs}| \geq \SI{0.5}{m/s}$.

Qualitative results are shown in Fig.~\ref{fig:qualitative_evaluation_positive_examples} and Fig.~\ref{fig:qualitative_evaluation_negative_examples}. The first graphic depicts the measurement that was already visualized in Fig.~\ref{fig:example_specular_reflection}. The performance for this example is very good, with a high ratio of clutter identified and not a single false positive detection, i.e. one that corresponds to a real moving object but is classified as clutter. It is worthwhile to note that the algorithm's internal method for finding clutter resulting from specular reflection (presented in Section~\ref{section:methods}) succeeds even though there are more clutter detections than correctly positioned ones. Another positive example is shown in Fig.~\ref{fig:qualitative_evaluation_positive_example_2}. It proves that the algorithm, although being primarily designed for extra-urban traffic, can also achieve considerable results in dense low velocity situations.
In contrast, two cases of failure are highlighted in Fig.~\ref{fig:qualitative_evaluation_negative_examples}. In the first image, several of the detections that correspond to the truck on the opposite lane are falsely predicted to be clutter. While two of these errors can be attributed to a negated velocity value, the remaining ones are caused by the heuristic used for identifying underbody reflections. The employed ruleset cannot differentiate between a small vehicle at which such reflections occur and a large object with detections scattered across its entire length. Fig.~\ref{fig:qualitative_evaluation_negative_example_2} shows a detection corresponding to a car in front of the sensor which is mistakenly classified as clutter. This, in turn, leads to one of the vehicle's specular reflections not being found as its distance is too large to fit to the only remaining detection at the car's position.

Overall during observation, it was noticed that false positives usually occur such that there are still some detections left which correspond to the same objects and are correctly classified (e.g. as in Fig.~\ref{fig:qualitative_evaluation_negative_example_1} and~\ref{fig:qualitative_evaluation_negative_example_2}). This is important for the detection of said objects by subsequent processing steps like object detection or tracking. Moreover, a large portion of the mentioned errors is located on the opposite lane behind a guardrail.

The results of quantitative evaluation are listed in Tab.~\ref{tab:performance}. Formulas for all metrics can be found e.g. in~\cite{Naser2020}. Only the algorithm's predictions for moving detections labeled as clutter or nonclutter are analyzed. This is because stationary detections cannot be clutter according to the definition in Section~\ref{section:clutter_effects} and are thus not classified by the algorithm. Ambiguous detections must be excluded since their true class isn't known.
It should be noted that the results would be drastically improved by incorporating the class for detections stemming from stationary reflection points. Since vehicles moving tangentially to the sensor are very rare in the data, it would almost always be correct to simply output this third class for detections with low velocity. However, the performance metrics' increase would derogate their informative value regarding the actual algorithm's effectiveness.

\addtolength{\textheight}{-2.2cm}

\begin{table}[!b]
	\centering
	\caption{Performance of the presented algorithm}
	\label{tab:performance}
	\begin{tabular}{ c | c | c | c | c  }
		Precision & Recall & Specificity & Balanced Accuracy & F1-Score\\
		\hline
		\rule{0pt}{8pt} \SI{98.47}{\percent} & \SI{79.86}{\percent} & \SI{86.03}{\percent} & \SI{82.95}{\percent} & \SI{88.20}{\percent}
	\end{tabular}
\end{table}
As can be seen in Tab.~\ref{tab:performance}, the presented algorithm achieves strong performance in all categories. Almost \SI{80}{\percent} of all clutter is detected, while only \SI{14}{\percent} of the detections corresponding to real moving objects are falsely classified. If a detection is predicted to be clutter, this decision is correct in more than \SI{98}{\percent} of cases.
The unbalance between recall and specificity is desired since false positives are significantly more severe errors than missing some clutter. The former can hinder effective detection of road users by subsequent processing steps and should thus be avoided when possible at reasonable cost. In comparison, some clutter detections going undetected poses less of a burden to the respective methods. After all, they would have to cope with a much higher amount of clutter if the algorithm for filtering it wasn't employed.

Execution of the clutter detection algorithm only takes about \SI{1.0}{ms} per sensor measurement on average on an AMD Ryzen 7 3700X CPU (without any parallelization and not counting the preceding search for reflection surfaces). This is of great importance because it means only minimal delay is introduced when using the algorithm as a preprocessing step before other methods.

Comparing the results to other publications (see Section~\ref{section:related_work}) is not possible. As none of the data sets used in those works was published, evaluating the presented algorithm on one of them is not an option.
For lack of publicly available code or knowledge about implementation details, replicating the approaches isn't feasible either. Finally, direct comparison of the stated performance values isn't suitable for several reasons.
In contrast to all related work, the data used in this paper reflects real extra-urban traffic. It includes several kinds of vehicles like cars and trucks, many highly reflective surfaces such as guardrails, and oncoming traffic. A typical mass-produced radar sensor with large FoV was used for recording. Thus, various types of clutter occur. Detections are processed immediately upon measurement and clutter of any kind has to be identified. In other works, however, depending on the exact publication, data was recorded in vastly different scenarios, a more powerful sensor was used or the considered classes differ significantly. Furthermore, none of the authors state their method's execution time. It is to be expected that every approach employing a neural network requires a multiple of the proposed algorithm's duration.

\section{Conclusion} \label{section:conclusion}

In this work, a rule-based algorithm is presented that classifies each moving detection in an automotive radar sensor's measurement either as clutter or as nonclutter. Unsystematic clutter is detected by simple filtering and a search for similar other detections. Three further checks each aim at identifying a specific type of clutter by analyzing whether the respective corresponding propagation path is likely to have occurred.

The algorithm is evaluated using a data set recorded in extra-urban traffic and labeled in a semi-automatic process. A quantitative analysis shows that almost \SI{80}{\percent} of all clutter is detected, while only less than \SI{14}{\percent} of the detections corresponding to real objects are mistakenly classified as clutter. Together with a qualitative inspection, this proves that the algorithm's concept of geometrically modeling propagation paths is well suited for identifying clutter. However, a systematic shortcoming of the heuristic used for detecting clutter caused by underbody reflections can be observed. The algorithm's execution time is very low, lying at only about \SI{1.0}{ms} per sensor measurement.

Future work could involve the incorporation of other effects causing clutter, provided a suitable test can be formulated. Alternatively, existing methods could be enhanced, e.g. by improving the mentioned heuristic or by also considering moving reflection surfaces.

\bibliographystyle{IEEEtran}
\bibliography{references}

\end{document}